\pdfoutput=1
\documentclass[11pt]{article}
\usepackage{emnlp2023}
\usepackage{times}
\usepackage{latexsym}
\usepackage[T1]{fontenc}
\usepackage[utf8]{inputenc}

\usepackage{amssymb}
\usepackage{adjustbox}
\usepackage{booktabs}
\usepackage{dirtytalk}
\usepackage{enumitem}
\usepackage{amsmath,graphicx}
\usepackage[noabbrev,capitalize,nameinlink]{cleveref}
\usepackage[shrink=30]{microtype}
\usepackage{inconsolata}
\usepackage{nicefrac}
\usepackage{tcolorbox}
\usepackage{xcolor}
\usepackage[T1]{fontenc}
\usepackage{amsmath}
\usepackage{multirow}

\crefname{equation}{equation}{equations}   
\crefname{footnote}{footnote}{footnotes}   
\crefname{section}{\S}{\S\S}
\Crefname{section}{\S}{\S\S}    
\crefformat{section}{#2\S#1#3}  
\Crefformat{section}{#2\S#1#3}
\crefrangeformat{section}{\S\S#3#1#4--#5#2#6}
\Crefrangeformat{section}{\S\S#3#1#4--#5#2#6}
\crefformat{section}{#2\S#1#3}  
\crefrangeformat{section}{\S\S#3#1#4--#5#2#6}

\newcommand{\baseline}[1]{\textsc{#1}}

\newcommand{\slt}{spoken language translation}

\providecommand{\citep}[1]{\cite{#1}}
\newcommand{\delim}{\textbf{<SENT>}}
\definecolor{primary}{HTML}{EA4335}
\definecolor{accent}{HTML}{C5221F}
\definecolor{background}{HTML}{F5F5F5}

\definecolor{error}{HTML}{C5221F}
\definecolor{correct}{HTML}{008566}

\usepackage{todonotes}
\title{

Long-Form Speech Translation through Segmentation with
\\ Finite-State Decoding Constraints on Large Language Models

}
\author{
Arya D. McCarthy$^{\star}$\textnormal{,} Hao Zhang$^{\dagger}$\textnormal{,} Shankar Kumar$^{\dagger}$\textnormal{,} Felix Stahlberg$^{\dagger}$\textnormal{, and} Ke Wu$^{\dagger}$ \\
\\
$^{\star}$Center for Language and Speech Processing, Johns Hopkins University \\
$^{\dagger}$Google Research\\
\texttt{mccarthy@jhu.edu} \\
\texttt{\{haozhang,shankarkumar,fstahlberg,wuke\}@google.com} 
}

\begin{document}
\maketitle
\begin{abstract}
One challenge in speech translation is that plenty of
spoken content is long-form, but short units are necessary for
obtaining high-quality translations. To address this mismatch,
we adapt large language models (LLMs) to split
long ASR transcripts into segments that can be independently
translated so as to maximize the overall translation quality.
We overcome the tendency of hallucination in LLMs by incorporating finite-state constraints during decoding; these eliminate invalid outputs without requiring additional training.
We discover that LLMs are adaptable to transcripts containing ASR errors through prompt-tuning or fine-tuning.
Relative to a state-of-the-art automatic punctuation baseline, our best LLM improves the average BLEU by 2.9~points for English--German, English--Spanish, and English--Arabic TED talk translation in 9~test sets, just by improving segmentation.


\end{abstract}

\section{Introduction}

With the proliferation of long-form audiovisual content online, translation and captioning become paramount for accessibility. Cascade models remain the dominant approach for speech translation \citep{arivazhagan-etal-2020-retranslation,li-etal-2021-sentence}, decomposing the problem into automatic speech recognition (ASR), post-processing of the transcript, and machine translation (MT).

The cascade's MT component typically operates on sentence-like units, with each sentence translated independently of the others. When asked to translate long passages, models regularly fail or degenerate \citep{cho-etal-2014-properties,pouget-abadie-etal-2014-overcoming,koehn-knowles-2017-six}.
This differs considerably from the expectations for automatic speech recognition models (e.g. \citealp{rnnt}) that can process inputs of unbounded lengths.
MT models must either be able to cope with potentially long, multi-sentence inputs or, alternatively, they must be able to determine cutpoints at which the transcript
can be segmented into compact, independently translatable units. This work introduces a new, effective approach for the latter.

While numerous text segmentation techniques have been proposed to improve \slt{} (\cref{sec:related-work}), 
the problem remains hard and unsolved.
Indeed, Li et al.~\citep{li-etal-2021-sentence} demonstrate that poor
sentence segmentation degrades performance almost twice as
much as transcript lexical errors. 

We cast sentence segmentation as a sequence-to-sequence task, rather than a traditional structured prediction task that tags sentence-final tokens.
While this lets us leverage large language models, such models' outputs can be ill-formed.
Even by using additional data for fine-tuning, residual adapters \citep{tomanek-etal-2021-residual,chronopoulou-etal-2022-efficient}, or future discriminators \citep{yang-klein-2021-fudge}, simple syntactic constraints can be difficult to enforce.
Moreover, all three require modifying the model or storing additional learned parameters.
In light of these concerns, we introduce a simple, flexible, and modular approach to generating well-formed task-specific strings at inference time without any additional training.
We compactly express constraints on the output format as finite-state machines, then efficiently enforce these via composition.
While the approach is simple, it remains unexplored for large language models, and it yields automatic gains on downstream performance, advancing the state of the art for speech translation and thereby applicable to existing systems.
Moreover, the approach is sufficiently general that it can be applied to other domains in a \emph{plug-and-play} manner. 



We benchmark our approach as a component in a speech translation cascade.
Experiments in three language
pairs indicate that our approach outperforms both a baseline 
cascade system that predicts
punctuation marks before inferring sentence boundaries and
a strong neural structured prediction model. Overall, we improve
the BLEU score on the IWSLT test sets by 2.9 points, closing~\(\nicefrac{3}{4}\) of the gap between the previous best and the oracle system.
Our contributions are three-fold:
\begin{enumerate}
    \item We propose a novel LLM-based approach for long-form speech translation, which can be applied to any ASR-MT speech translation cascade system and yield a significant increase in translation quality.
    \item To the best of our knowledge, we are the first to investigate the use of finite-state decoding constraints in combination with LLMs to produce consistent improvements.
    \item We report additional small but consistent improvements by prompt-tuning or fine-tuning LLMs on ASR transcripts containing lexical and grammatical errors.
\end{enumerate}

\section{Windowing Approach}
\label{sec:windowing}

\begin{figure*}[htb]
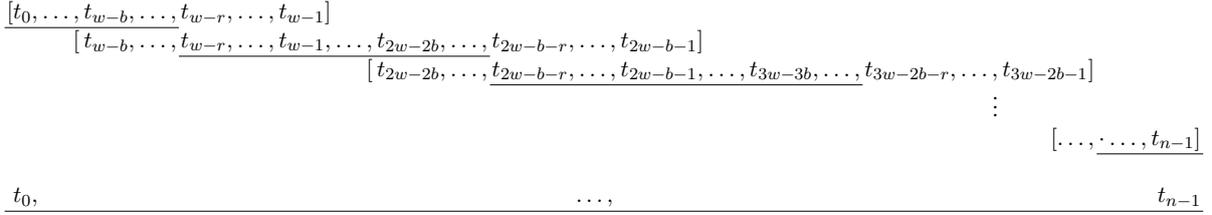

  \centering
\(
\resizebox{\textwidth}{!}{
\arraycolsep=1.0pt\def\arraystretch{1.0}
$
\begin{array}{rrrrrrrrrrrrrrrrrrrrrr}
 [t_0, & \ldots, & t_{w-b}, & \ldots, & t_{w-r}, & \ldots, & t_{w-1}] & & & & & & & & & & & & &  & \\
 \cline{1-4}
        &        [& t_{w-b}, & \ldots, & t_{w-r}, & \ldots, & t_{w-1}, & \ldots, & t_{2w-2b}, & \ldots, & t_{2w-b-r}, & \ldots, & t_{2w-b-1}] & & & & & & &  &\\
                                      \cline{5-10}
        &         &         &         &         &         &         &       [ & t_{2w-2b}, & \ldots, & t_{2w-b-r}, & \ldots, & t_{2w-b-1}, &\ldots, & t_{3w-3b}, & \ldots, & t_{3w-2b-r}, & \ldots, & t_{3w-2b-1} ]   & \\
\cline{11-16}
        &         &         &         &         &         &         &         &          &         &          &         &          &        &         &          &          & \vdots  &             & & &\\                                                      &         &         &         &         &         &         &         &          &         &          &         &          &        &         &          &          &         & [\ldots, & \cdot & \ldots, & t_{n-1}]    \\
\cline{20-22} \\ 
\hphantom{[}t_0, & \multicolumn{20}{c}{\ldots,} & t_{n-1}\\                                   
\cline{1-22}                                                         
\end{array}
$
}
\)
\caption{Processing overlapping windows instead of entire transcript passages. $w$ is the window size used in both training and inference. $b$ is the total context window size. $r$ ($\le b$) is the right context window size. The underlines below the windows indicate which local segmentation decisions are taken as global decisions. Portions not underlined (i.e., the context window) are still provided to the segmentation model to inform segmentation of underlined portions. 
}
  \label{fig:windowing}
\end{figure*}

One major challenge in modeling and inference of long-form transcript segmentation is that the input sequences can be very long. For example, a TED talk can contain more than one thousand words ~\citep{li-etal-2021-sentence}.
We take a divide-and-conquer approach that operationalizes two straightforward principles in modeling.
First, words on the left and right are both useful for deciding if a sentence delimiter should be present at the current word position.
Second, distant words are less useful than nearby words. From these two principles, we design a top-level sliding window algorithm to balance the need for bidirectional modeling and efficiency of computation.
We divide the passage into windows at both training and test time, with a small context window on each side to inform decisions at window edges (\cref{fig:windowing}).
With this top-level inference algorithm, the sequence-to-sequence machine learning problem is now reduced to the window-level. 
The problem is now to predict a sequence of segmentation decisions $\mathbf{y}=y_1,\ldots,y_{w}$ for each text \emph{window} of size at most $w$ tokens: $\mathbf{x}=x_1,\ldots,x_{w}$. 

\section{Modeling Approaches}

A classic approach to discriminative sequence modeling is the conditional random field (CRF) \citep{lafferty-etal-2001-conditional,liu-etal-2005-using-conditional}. This conditional graphical model allows incorporating arbitrary features of the transcript, including linguistic variables and word embeddings. 

\subsection{Structured Prediction Baseline: Bidirectional RNN Model}
\label{sec:birnn_model}
The limitation of the CRF is in the Markov assumption it makes, considering only the immediately previous word's segmentation decision.
Even higher-order CRFs can only consider a fixed-size history within $\mathbf{y}$. 
Instead, we introduce a neural autoregressive segmenter. It is an encoder--decoder neural network with monotonic hard attention to the bidirectionally encoded input at the current word position, admitting the same rich featurization of $\mathbf{x}$ as the CRF; its likelihood is
\begin{align}
  p_\theta(\mathbf{y} \mid \mathbf{x}) =& \prod_{t=1}^w p_\theta(y_{t} \mid \mathbf{y}_{<t}, \mathbf{x})  \\
                                      :=& \prod_{t=1}^w p_\theta\bigl(y_{t} \mid \mathbf{y}_{<t}, \mathbf{BiRNN(x)}_t\bigr)
\end{align}
where $p_\theta$ is parameterized by a recurrent neural network followed by a linear projection layer and a softmax to obtain a locally normalized distribution.
Exact inference here is intractable (unlike a CRF); we approximate it with beam search.
This model and a QRNN-based  \cite{DBLP:conf/iclr/0002MXS17} automatic punctuation model will serve as baselines.

\subsection{Large Language Models for Segmentation}

More recently, the paradigm of pre-training followed by fine-tuning or few-shot learning has achieved great successes across many NLP tasks. The pre-training task is typically  a variant of a language model \cite{gpt,palm} or an autoencoder \cite{t5} where a corrupted version of a sentence is mapped to its uncorrupted counterpart. We can encode segmentation as such a task: reproducing the input with inserted sentence delimiters.
Concretely, we encode $\mathbf{y}$ as $z_1,\ldots,z_{w}$ where $z_t = \operatorname{Concat}(d_t,x_t)$ and $d_t \in \{\epsilon, \blacksquare\}$.
For example, we feed \texttt{i am hungry i am sleepy} to the model, and it produces the sentence-delimited string \texttt{i am hungry $\blacksquare$ i am sleepy}.
We use the publicly available T5 (Text-to-Text Transfer Transformer) model \citep{t5} and the GPT-style \cite{gpt} PaLM model \cite{palm} as the foundation for our text-based segmenters.  

\subsubsection{Prompting and Fine-tuning}
\begin{figure}[t]
    \centering
    \begin{tcolorbox}[left=1mm,right=1mm,top=1mm,bottom=1mm,middle=1mm,colback=background]
    \scriptsize \color{gray}

    \textcolor{black}{>{}>{}>} Segment a sequence of words into sentences separated by the delimiter \delim{}
\\\\
Input: well first of all thank you so much that is a beautiful compliment i do think he is the best interviewer alive\\
Segmented Output: well first of all thank you so much \delim{} that is a beautiful 
 compliment \delim{} i do think he is the best interviewer alive
\\\\
Input: i remember when my dad had to leave our home in scranton pennsylvania to find work i grew up in a family where if the price of food went up you felt it that's why one of the first things i did as president was fight to pass the american rescue plan\\
Segmented Output: i remember when my dad had to leave our home in scranton pennsylvania to find work  \delim{} i grew up in a family where if the price of food went up you felt it  \delim{} that's why one of the first things i did as president was fight to pass the american rescue plan
\\\\
Input: we're done talking about infrastructure weeks we're going to have an infrastructure decade\\
Segmented Output:  we're done talking about infrastructure weeks \delim{} we're going to have an infrastructure decade

    \tcblower
\scriptsize \color{gray}

Input: \textcolor{accent}{it is going to rain today remember to bring an umbrella}\\
Segmented Output:

    \end{tcolorbox}

    \caption{Prompting PaLM to segment a text window (red) based on three examples.}
    \label{fig:visual-abstract}
\end{figure}

Training examples for this task look like the input output pairs in \cref{fig:visual-abstract}. In fine-tuning, we update the full set of parameters for a given model on such examples to minimize the cross entropy on the output. For T5 models, the input sequence will be fed to the encoder, and the output sequence will be fed to the decoder through teacher forcing. For PaLM models, the input sequence and the output sequence are concatenated and fed to the decoder with an optional prompt as the prefix.
For decoder-only PaLM models, a text prompt like the one in \cref{fig:visual-abstract} or a fine-tuned soft prompt \cite{lester-etal-2021-power} in the embedding space prompts the decoder to enter the state for the segmentation task. When we fine-tune PaLM, the entire model is updated for this task so that no prompting is necessary.

\subsubsection{Decoding Constraints}
A deficiency of generation with LLM 
is that the output might not only fail to correctly segment the passage; it might not even contain the same tokens as the passage. We shall say that an output is \emph{well-formed} if it contains the same token sequence as the input, with zero or one sentence delimiters before each token.
While the rich parameterization of such large Transformer models might \emph{learn} the inherent structure of the output, we provide two solutions to \emph{enforce} well-formedness. 

Both approaches share the attractive quality of being \emph{plug-and-play}: they require no additional parameter-learning, and they can be coupled with an already-trained language model.




\paragraph{Levenshtein Alignment for Post-processing} \label{sec:levenshtein}
\label{sec:levenshtein_alignment}
The generation models' ability to produce arbitrary outputs may be seen as a strength: the model could correct transcription errors and remove disfluencies, if so trained.
Therefore, we can let the model generate freely without enforcing structural constraints, then enforce well-formedness post-hoc.
\citet{kumar02_icslp} describe a WFST for \textit{Levenshtein alignment} between two strings. We use it to align the generated string with $\mathbf{x}$. We then project segment boundaries across alignment links from the generated string onto $\mathbf{x}$ to determine $\mathbf{y}$. In this way, annotations can be salvaged when LLM does not precisely recreate the input.

\paragraph{Finite-State Constraints in Decoding}
\label{sec:fst_constraint}

\begin{figure*}[htb]
    \centering
    \includegraphics[width=\linewidth]{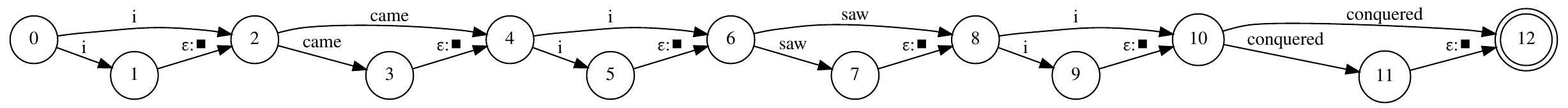}
    \caption{FST representing all possible segmentations for the transcript \protect\say{i came i saw i conquered}.
      }
    \label{fig:sawtooth}
\end{figure*}

A natural strategy to force well-formed outputs is \emph{constrained decoding} (e.g.\ \citealp{zhang-etal-2019-neural}).
In it, we compose the input FSA $\mathbf{x}$ and a special FST $\mathcal{T}$ encoding all possible segmentation decisions, then project the FST to the output tape to obtain a determinized FSA for the output space. The FST $\mathbf{x} \circ \mathcal{T}$ is shown in \cref{fig:sawtooth}.


An advantage of the finite-state approach is that \emph{any} constraint expressible as a regular language is possible. Consequently, our implemented system is applicable a large class of tagging and parsing problems in NLP, not just sentence segmentation. For instance, NP chunking \citep{ramshaw-marcus-1995-text} and BIO tagging, truecasing \citep{lita-etal-2003-truecasing}, retokenization, tetra-tagging for syntactic parsing \citep{kitaev-klein-2020-tetra}, and lexically constrained decoding \citep{hasler-etal-2018-neural} can all be framed as finite-state transformations of an input sequence.

\section{Experiments}

We evaluate our proposed method for using large language models for long-form speech translation with three sets of experiments: (1) analysis of hyperparameters, (2) comparison with competing methods, and (3) robustness to speech recognition errors.
In each case, we are concerned with translation quality as measured by BLEU.
We also assess the LLM output directly by qualitative analysis, well-formedness percentage, and (for diagnostic purposes, following \citealp{Goldwater:2009aa}) segmentation \(F_1\) score against the sentence-segmented reference.

Our experiments are carried out on the IWSLT speech translation data sets, subjected to the same pre-processing as described in \citet{li-etal-2021-sentence}. We use the 2014 data for dev and 2015 and 2018 for test.
The fourteen reference transcripts in our dev set range from 861 to 1234 words; by contrast, the median length of a sentence in written English is close to 17 words \citep{kucera1970computational}. 
We use the publicly available Speech-to-Text Google API%
\footnote{%
    \url{https://cloud.google.com/speech-to-text}
} 
to generate ASR transcripts.
We remove the automatically predicted punctuation and lowercase the ASR transcripts and use English--\{German,Spanish,Arabic\} MT models trained with the same preprocessing on the source side as~\citealp{li-etal-2021-sentence}.
The MT model is a Transformer with a model dimension of 1024, hidden size of 8192, 16 attention heads, 6 encoder layers, and 8 decoder layers. We decode with a beam size of 4. 
In our experiments, the three MT model instances and the ASR model (and thereby its transcripts) are fixed while we vary the sentence segmentation policies.




\subsection{Context Window Size}
\begin{figure}
    \centering
    \includegraphics[width=\columnwidth]{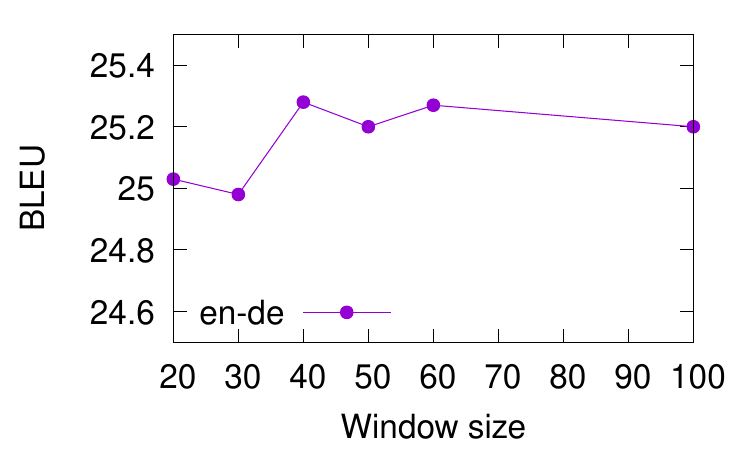}
    \caption{BLEU for English--German as context window size for segmentation increases. Each dot represents a T5 segmentation model trained with the same window size for inference time.}
    \label{fig:window-bleu}
\end{figure}

In \cref{sec:windowing}, we introduced the top-level sliding window inference algorithm above all modeling choices. To compare different models fairly, we fix the hyperparameters $(w,b,r)=(40,10,5)$ for the algorithm throughout the experiments.  This choice is guided by a linear search over the window lengths $w$ in the range of $[20, 100]$. The overlapping buffer size for both ends is set to 5 based on findings of segmentation for \emph{punctuated} text \citep{wicks-post-2021-unified}. According to \cref{fig:window-bleu}, translation quality degrades slightly as window size approaches 20. But very large windows do not appear to be beneficial. The observation validates the two guiding principles of our sliding window approach.

\subsection{Choice of Prompt}
\begin{figure}
    \centering
    \includegraphics[width=\columnwidth]{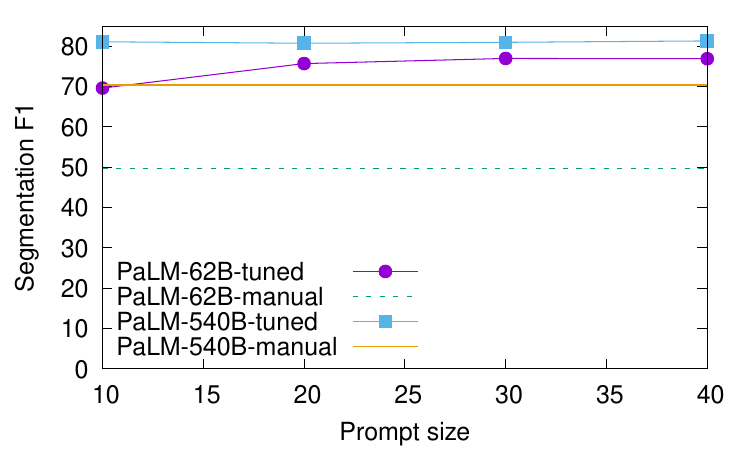}
    \caption{Segmentation \(F_1\) on the dev set as prompt size varies.}
    \label{fig:prompt-f1}
\end{figure}

The manual prompt in \cref{fig:visual-abstract} is the one we selected from a few variants for the decoder-only PaLM models. Instead of exploring the unbounded space of prompts, we resorted to the more principled method of prompt tuning \cite{lester-etal-2021-power} to optimize the prompt in the embedding space for the segmentation task.
For prompt tuning, the only hyperparameter is the length of the embedding prompt (the embedding size is tied to the corresponding model). In \cref{fig:prompt-f1}, we show that for the PaLM models of 62B and 540B, an embedded prompt as short as 10 can achieve much higher \(F_1\) than our hand-written prompt.
But it is also notable that the gap between prompt tuning and manual prompting shrank from 25 percent to 10 percent as the model size increased from 62B to 540B, indicating the increasingly stronger generalization capability of extremely large language models. Based on \cref{fig:prompt-f1}, we use 30-token soft prompts in the main results.

\subsection{Effect of Finite-State Constraints}
We make a contrast between greedy search and beam search, with either the segmentation FST constraint~\cref{sec:fst_constraint} inside the decoder or post-hoc Levenshtein alignment~\cref{sec:levenshtein} for repairing invalid output. We also vary the model types and model sizes to analyze the impact of constrained decoding in different situations.
\cref{tab:search_ablation} shows that constraints are crucial for smaller models in prompt-tuned scenarios. For example, the rate of output being well-formed is only 14.5\% using greedy search for the PaLM 8B model. Even when the model size is increased to 62B, the wellformedness rate is still below 90\%. The Levenshtein post-alignment algorithm is effective. But the more general finite-state constraint is even more effective. For the 8B model, the improvement in \(F_1\) is 1--2\% absolute. For the 62B model, the improvement is nearly 3\% absolute. On the other hand, if the cost of fine-tuning is acceptable, LLMs can adapt to this task very well. The fine-tuned T5 base model has a wellformedness rate of 99.4\% (the rate is even higher for the T5 11B model: 99.8\%). But we shall point out that for the results to be useful to downstream applications, either of the two types of constraints is necessary to completely eliminate hallucinations from LLMs. And the FST constraints are more general and more effective as they affect beam search by rejecting non-wellformed hypotheses during search.

\begin{table}[htb]
  \center{
  \begin{adjustbox}{max width=\linewidth}
  \begin{tabular}{@{}l| ll rr @{}}
    \toprule
     \baseline{model} & \baseline{constraint} &\baseline{search} &\textsc{wellformed} & \textsc{F1}\\
    \midrule
    \multirow{5}{*}{\baseline{T5 base}}&\baseline{unconstrained} &\baseline{greedy} & 99.4\% &  -- \\
    \multirow{6}{*}{\baseline{Fine Tuned}}& {} &\baseline{beam=4} & 99.4\% &  -- \\
    \addlinespace    
    &\baseline{Levenshtein}   &\baseline{greedy} & 100.0\% & 0.786\\
    &                      {} &\baseline{beam=4} & 100.0\% & {\bf 0.788} \\
    \addlinespace    
    &\baseline{FST}           &\baseline{greedy} & 100.0\% & 0.786\\
    &                      {} &\baseline{beam=4} & 100.0\% & {\bf 0.788} \\
    \midrule
    \midrule
    \multirow{5}{*}{\baseline{PaLM 8B}}&\baseline{unconstrained} &\baseline{greedy} & 14.5\% &  -- \\
    \multirow{6}{*}{\baseline{Prompt Tuned}}&                      {} &\baseline{beam=4} & 52.7\% &  -- \\
    \addlinespace    
    &\baseline{Levenshtein}   &\baseline{greedy} & 100.0\% & 0.715\\
    &                      {} &\baseline{beam=4} & 100.0\% & 0.689 \\
    \addlinespace    
    &\baseline{FST}           &\baseline{greedy} & 100.0\% & 0.717 \\
    &                      {} &\baseline{beam=4} & 100.0\% & {\bf 0.727} \\
    \midrule
    \multirow{5}{*}{\baseline{PaLM 62B}}&\baseline{unconstrained} &\baseline{greedy} & 85.9\% &  -- \\
    \multirow{6}{*}{\baseline{Prompt Tuned}}&                      {} &\baseline{beam=4} & 89.0\% &  -- \\
    \addlinespace    
    &\baseline{Levenshtein}   &\baseline{greedy} & 100.0\% & 0.735\\
    &                      {} &\baseline{beam=4} & 100.0\% & 0.737 \\
    \addlinespace    
    &\baseline{FST}           &\baseline{greedy} & 100.0\% & 0.761 \\
    &                      {} &\baseline{beam=4} & 100.0\% & {\bf 0.764} \\
    \bottomrule
  \end{tabular}
  \end{adjustbox}
  }
  \caption{Effect of finite-state decoding constraints and Levenshtein post alignment on segmentation \(F_1\).}
  \label{tab:search_ablation}  
\end{table}

\subsection{Main Results: LLMs against Structured Prediction Models}

\begin{table*}[htb]
    \centering
    \begin{adjustbox}{max width=\linewidth}
    \begin{tabular}{@{} l r r r r r r r r r r r @{}}
    \toprule
     &F1 & \multicolumn{3}{c}{\textsc{en--de}} & \multicolumn{3}{c}{\textsc{en--es}} & \multicolumn{3}{c}{\textsc{en--ar}} & \\
     \cmidrule(lr){3-5} \cmidrule(lr){6-8} \cmidrule(lr){9-11}
     Policy  & TED 2014 &  2014 & 2015 & 2018 &  2014 & 2015 & 2018 &  2014 & 2015 & 2018 & Avg\\
     \midrule
    \textit{Baselines and Oracle:}\\
     \quad\baseline{Oracle}        & 1.000  & 26.66  & 30.24  & 25.21  & 40.38  & 41.72  & 41.84 & 15.66 & 18.18 & 17.59 & 29.62\\
     \quad\baseline{FixedLength}   & 0.041  & 20.82  & 23.45  & 19.66  & 32.76  & 34.03  & 34.01 & 12.64 & 14.79 & 13.92 & 23.66\\
        \quad\textsc{\citet{li-etal-2021-sentence}} & -- & -- & 27.00 & 22.00 & -- & -- & -- & -- & -- & -- & --\\

    \addlinespace
    \textit{Small Structured Prediction Models:}\\
     \quad\baseline{Punctuate}     & --    & 22.80  & 26.30  & 21.60  & 35.70  & 36.90  & 36.70 & 13.70 & 15.80 & 15.40 & 25.81\\     

    \quad\baseline{BiGRU f.t.}     & 0.697  & 24.55  & 28.10  & 23.14  & 37.31  & 39.08  & 38.64 & 14.41 & 16.77 & 16.19 & 27.39 \\
    \addlinespace
    \textit{LLMs:}\\
    \quad\baseline{T5-base}        & 0.788  & 25.28 & 29.14   & 24.05    & 38.75  & 40.23  & 39.96   & 14.94 & 17.32  & 16.57 & 28.33 \\
    \quad\baseline{T5-11B}         & 0.821  & 25.63 & \textbf{29.63}   & \textbf{24.27}    & \textbf{39.16}  & 40.64  & \textbf{40.05}   & \textbf{15.31} & 17.60  & 16.48 & \textbf{28.66} \\
    \quad\baseline{T5-11B-ASR}         & \textbf{0.836}  & 25.71 & 29.28   & 24.22   & 39.11  & 40.47  & 40.02   & 15.24 & 17.58  & 16.66 & 28.59\\    
    
   \quad\baseline{PaLM-PromptTuned-62B}        & 0.764 & 25.10 & 28.69  & 23.92  & 38.52 &  40.01& 39.22 & 15.03 & 17.13  &  16.58 & 28.08 \\
   \quad\baseline{PaLM-PromptTuned-62B-ASR}     & 0.781  & 25.15 & 29.09 & 23.71 & 38.69 & 40.07 & 39.31 &15.13  &17.21 & 16.76 & 28.17\\
   \quad\baseline{PaLM-FineTuned-62B}     & 0.820  & 25.71  & 29.19  & 23.97 & 38.96  & 40.56 & 39.74 & 15.07  & 17.66 &  \textbf{16.90} & 28.51\\   
   \quad\baseline{PaLM-FineTuned-62B-ASR}     & 0.832  & \textbf{25.84} & 29.37  & 24.13 & 39.02 &40.46 & 39.89 & 15.17 &\textbf{17.80} & 16.65 & 28.61 \\
      \quad\baseline{PaLM-PromptTuned-540B}     & 0.816 & 25.44 & 29.29 & 24.23 & 38.95 & \textbf{40.70} & 39.74 & 15.03 & 17.61 & 
      16.86 & 28.49 \\
   \quad\baseline{PaLM-PromptTuned-540B-ASR}     & 0.835  & 25.52 & 29.37 &24.15 & 39.08  & 40.67 & 39.98 & 15.11 & 17.64 & 16.61 & 28.56  \\   

    \bottomrule
    \end{tabular}
    \end{adjustbox}
  \caption{Segmentation F1 scores on dev set and BLEU scores on dev and test sets, translating into German, Spanish, and Arabic.
  }
    \label{tab:results}
    
\end{table*}

Using the IWSLT TED datasets as preprocessed by \citet{li-etal-2021-sentence}, we compare LLM models against their approach, two strong custom structured prediction baselines. We also report the performance of an oracle segmenter.

\begin{description}
    \item[\baseline{FixedLength}] 
    Separates the transcript into disjoint segments with the same number of tokens. While this requires no external segmentation model, the resulting segments are non-sentential \citep{tsiamas2022shas}.
    \item[\baseline{Oracle}] 
    Uses punctuation from the reference transcripts to segment. The segmentation is projected onto Levenshtein-aligned words in the noisy ASR transcripts (\cref{sec:levenshtein}).\footnote{A true oracle would optimize corpus-level BLEU over all $2^n$ segmentations, but this is intractable.}
    \item[\baseline{Punctuate}] 
    An interpretable two-pass segmentation that first infers punctuation \citep{soboleva2021replacing}, then uses a fixed set of inference rules to differentiate sentence-terminal punctuation marks from sentence-internal ones as in “St. John” and “The end.”

\item[\baseline{biGRU f.t.}]
  On the IWSTL data, fine-tunes a shallow biGRU model (\cref{sec:birnn_model}) trained on the C4 data set \cite{t5} using the same rules in \baseline{Punctuate} to derive sentence boundaries as supervision. The model has 1 left-to-right GRU layer, 1 right-to-left GRU layer, and 1 GRU layer in the decoder. It uses embeddings of character $n$-gram projections \citep{zhang-etal-2019-neural}.
\item[\baseline {T5-\{base,11B\}}]
  Fine-tunes the base or 11B (xxl) T5 model~\citep{t5} on the IWSLT data.
\item[\baseline {T5-11B-ASR}]
  Fine-tunes the 11B T5 model on the ASR output of IWSLT train and dev set. Sentence boundaries are projected from reference transcripts in the same way as \baseline{Oracle}.
\item[\baseline{PaLM-PromptTuned-\{62B,540B\}\{,-ASR\}}]
  Prompt-tunes the PaLM model \citep{palm} on the IWSLT data.
\item[\baseline{PaLM-FineTuned-62B\{,-ASR\}}]
  Fine-tunes the 62B PaLM model.

  \end{description}

The peer-reviewed state of the art for long-form speech translation is \citet{li-etal-2021-sentence} on the IWSLT data set for \textsc{en--de}. Compared to \baseline{Oracle}, there is still a large gap of 3 BLEU points which can be closed by improving segmentation alone.

\Cref{tab:results} lists the complete set of results.
\baseline{BiGRU F.T.} already beats \citet{li-etal-2021-sentence} by more than 1 BLEU point for \textsc{en--de}, proving itself as a strong structured prediction baseline. T5 and PaLM models improve the results furthermore. Within the T5 group, \baseline{T5-11B} improves over \baseline{T5-base} by 3\% in segmentation \(F_1\) which translates to consistent BLEU score improvement in almost all data sets. Within the PaLM group, the prompt-tuned 540B model is about 5\% more accurate than the 62B counterpart. Given the large number of parameters, fine-tuning PaLM models is very expensive. For the completeness of comparison, we include the fine-tune result for PaLM 62B. Its result is on-par with the T5 11B model. This fact  indicates that T5's encoder--decoder architecture has an inductive bias advantage over the PaLM model's decoder only architecture for this task, from a parameter efficiency point of view. But the strength of the PaLM family lies in its largest member. The 540B model with a tiny tuned prompt is as effective as the fully fine-tuned T5 11B or PaLM 62B.


\subsection{Robustness to Speech Recognition Errors}

One key difference between cascade speech translation and typical document-level translation is that transcription errors can be introduced, which propagate into the translation. 
When the input to segmentation models contains speech recognition errors, can such models still predict sentence boundaries accurately? The answer is yes, to a certain extent.
To test this, we replace the tuning data from ground-truth transcripts with punctuation-derived sentence boundaries to ASR transcripts that have sentence boundaries projected from their parallel ground-truth transcript counterparts. For example, we will tune the models to predict the segmentation for the passage: \textit{this train leaves at for \delim{} the next train will arrive in ten minutes}, even though there is a lexical error (\textit{for} versus \textit{four}).

\begin{figure}
    \centering
    \includegraphics[width=\columnwidth]{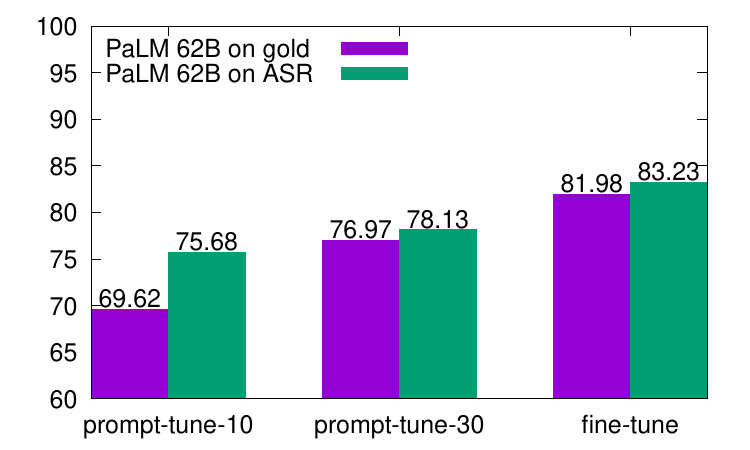}
    \caption{Contrast of segmentation \(F_1\) on the dev set between models trained on gold and ASR transcripts. }
    \label{fig:gold-vs-asr}
\end{figure}

\cref{tab:results} shows that training on the ASR transcripts is indeed beneficial.  On top of the strong results of the T5 11B model trained on ground-truth transcripts, the ASR version obtains another 1\% \(F_1\) improvement. The same is true for the PaLM 62B prompt-tuned and fine-tuned models.  The relative improvement is consistent across different prompt sizes and fine-tuning methods (\cref{fig:gold-vs-asr}). Still, the small segmentation improvement does not translate into significant BLEU score improvements.

\section{Error Analysis}
\subsection{Segment Length Histogram Analysis}

\begin{figure}
    \centering
    \includegraphics[width=\columnwidth]{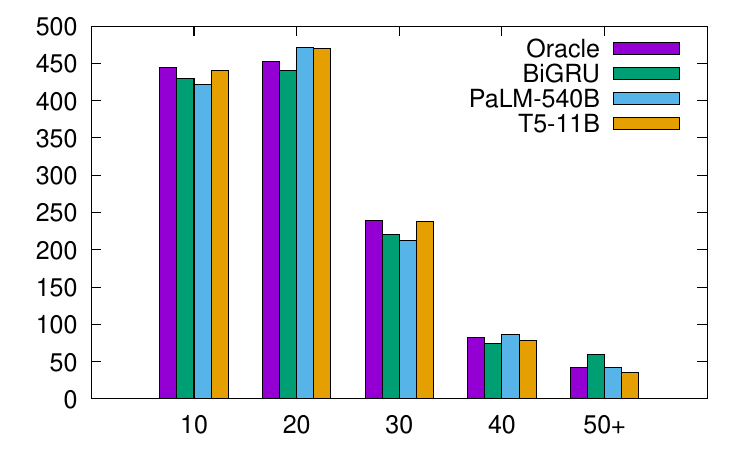}
    \caption{Histograms of segment lengths for \baseline{Oracle}, \baseline{BiGRU}, \baseline{PaLM PromptTuned 540B ASR}, and \baseline{T5 11B ASR}.}
    \label{fig:length-histo}
\end{figure}

To understand the improvements and the remaining errors, we first compare the length distribution of the \baseline{oracle}, the small model \baseline{BiGRU}, \baseline{T5-11B}, and \baseline{PaLM-540B}. \cref{fig:length-histo} indicates that the more very long ($\ge 50$) segments  a model has, the lower its \(F_1\) and BLEU scores tend to be. Both LLM models were able to reduce the number of very long segments, bringing it closer to the oracle.

\subsection{Qualitative Analysis}
\cref{tab:good-outputs} shows examples where the \baseline{T5-11B-ASR} model outperforms competing models. In the first two examples, the LLM model is able to capture the larger context and therefore make the correct prediction. The third example typifies the cases where \baseline{T5-11B}, which is fine-tuned on ground-truth transcripts without ASR errors, tends to make more wrong predictions when the input text is not fluent.

\cref{tab:bad-outputs} shows typical errors the \baseline{T5-11B-ASR} model makes. In the first two, ASR errors make the transcript difficult to parse. The third one is linguistically ambiguous. In the last one, the model's prediction is actually closer to the ground-truth segmentation than the Levenshtein-(mis)aligned ASR transcript.

Overall, LLMs such as \baseline{T5-11B-ASR} made real progress in predictions requiring longer context. However, even though fine-tuning on ASR transcripts improved robustness to disfluent input, overcoming ASR errors remains challenging.


\begin{table*}[t]
    \centering
    \begin{adjustbox}{max width=\textwidth}
\begin{tabular}{@{} l @{}} 
    \toprule
                    \textbf{Reference}:  this great renaissance for ancient egyptian art architecture and \textcolor{correct}{religion<SENT>egyptologists} have always known the site \\
        \textbf{ASR}: this great renaissance for ancient egyptian art architecture and \textcolor{correct}{religion<SENT>egyptologists} have always known the site\\
        \textbf{\baseline{BiGRU}}: this great renaissance for ancient egyptian art architecture and \textcolor{error}{religion egyptologists} have always known the site \\
        \textbf{\baseline{T5-11B-ASR}}:this great renaissance for ancient egyptian art architecture and \textcolor{correct}{religion<SENT>egyptologists} have always known the site\\
        \midrule
        \textbf{Reference}:  looking for layers of human \textcolor{correct}{occupation<SENT>and five meters down underneath a thick layer of mud we found} a dense layer of pottery\\
        \textbf{ASR}: looking for layers of human \textcolor{correct}{occupation<SENT>and five meters down underneath a thick layer of mud we found} a dense layer of pottery\\
        \textbf{\baseline{BiGRU}}: looking for layers of human \textcolor{error}{occupation and five meters down underneath a thick layer of mud<SENT>we found} a dense layer of pottery\\
        \textbf{\baseline{T5-11B-ASR}}: looking for layers of human \textcolor{correct}{occupation<SENT>and five meters down underneath a thick layer of mud we found} a dense layer of pottery\\

        \midrule
        \textbf{Reference}:\textcolor{correct}{actually started} in 1984 bc at a not-lost-for-long city found from above\\
        \textbf{ASR}: \textcolor{correct}{how actually actually started} in 1984 bc at a not lost for long city found from above\\
        \textbf{T5-11B}: \textcolor{error}{how<SENT>actually actually started} in 1984 bc at a not lost for long city found from above\\
        \textbf{T5-11B-ASR}: \textcolor{correct}{how actually actually started} in 1984 bc at a not lost for long city found from above\\
	\bottomrule
			\end{tabular}
			\end{adjustbox}
    \caption{Cases where the \baseline{T5-11B-ASR} model is more accurate.}
    \label{tab:good-outputs}
\end{table*}

\begin{table*}[t]
    \centering
    \begin{adjustbox}{max width=\textwidth}
\begin{tabular}{@{} l @{}} 
    \toprule
	\textbf{Reference}: designers can materialize their ideas directly in \textcolor{correct}{3d and surgeons} can practice on virtual organs underneath the screen  \\
 	\textbf{ASR}: designers can materialize their ideas directly in \textcolor{correct}{3d sturgeons} can practice a virtual audience underneath the screen\\
	\textbf{\baseline{T5-11B-ASR}}: designers can materialize their idesas directly in \textcolor{error}{3d<SENT>sturgeons} can practice a virtual audience underneath the screen.\\
	\midrule
 	\textbf{Reference}:  But our two hands still remain outside the \textcolor{correct}{screen <SENT> how} can you reach inside and interact with the digital information\\
	\textbf{ASR}: what are two hands still we made outside the screen \textcolor{correct}{that <SENT> how} can you reach inside and interact with the digital information\\
	\textbf{\baseline{T5-11B-ASR}}: what are two hands still we made outside the screen  \textcolor{error}{that how} can you reach inside and interact with the digital information\\
        \midrule
                \textbf{Reference}: this is really what brought me to using satellite \textcolor{correct}{imagery<SENT>for trying to map the past i knew} that i had to see differently\\
        \textbf{ASR}: this is really what brought me to using satellite \textcolor{correct}{imagery<SENT>for trying to map the past i knew} that i had to see differently\\
        \textbf{\baseline{T5-11B-ASR}}: this is really what brought me to using satellite \textcolor{error}{imagery for trying to map the past<SENT>i knew} that i had to see differently\\
        \midrule
                \textbf{Reference}: the equivalent of locating a needle in a haystack blindfolded wearing \textcolor{correct}{baseball mitts<SENT>so} what we did i\\
        \textbf{ASR}: the equivalent of locating a needle in a haystack blindfolded wearing \textcolor{error}{baseball<SENT>minutes so} what\\
        \textbf{\baseline{T5-11B-ASR}}: the equivalent of locating a needle in a haystack blindfolded wearing \textcolor{correct}{baseball minutes<SENT>so} what we did is\\
	\bottomrule
			\end{tabular}
			\end{adjustbox}
    \caption{Cases where the \baseline{T5-11B-ASR} model's prediction is wrong.}
    \label{tab:bad-outputs}
\end{table*}
\section{Related Work}
\label{sec:related-work}
\paragraph{Speech translation.}
While end-to-end systems for speech translation have exceeded the performance of cascade models on short sequences \citep{weiss2017sequence} even on public data \citep{9053406}, long-form audio is typically translated with cascades.
Previous work uses tagging approaches to separate text into independently translatable units.
Segmenting long texts into units suitable for translation has been a recurring topic in MT research \citep{li-etal-2021-sentence,8713737,pouget-abadie-etal-2014-overcoming,doi-sumita-2003-input,Goh2011SplittingLI}.
To bridge the gap between ASR and MT, \citet{li-etal-2021-sentence} address long-form speech translation. Claiming that segmentation is the bottleneck, they adapt their MT model to work \emph{with} automatic segmentations, however inaccurate they may be.

We are training our models to minimize the loss of source sentence segmentation. The ultimate objective is improving the downstream translation quality. It is interesting to explore reinforcement learning for segmentation \citep{srinivasan-dyer-2021-better}, but the state space is vast for the long-form segmentation problem compared to prior work on RL-based segmentation.

Finally, one may consider additional sources of data or training examples to improve modeling.
Using prosodic features when they are available is viable \citep{tsiamas2022shas}; however, we show that LLMs close most of the accuracy gap without these. As a contrasting approach, \citet{kumar02_icslp} focus on segmenting an ASR \textit{lattice}, rather than the decoded transcript. Finally, data augmentation \citep{pino-etal-2019-harnessing,9053406,li-etal-2021-sentence} can complement our approach. 

\paragraph{Text normalization and segmentation.}
\citet{mansfield-etal-2019-neural} model text normalization as a sequence-to-sequence problem, using \texttt{<self>} tags to bias toward copying, but they place no search constraints to ensure well-formedness. \citet{zhang-etal-2019-neural} also use finite automata intersected with a neurally generated lattice during decoding.

\citet{wicks-post-2021-unified} provide a unified solution for segmenting punctuated text in many languages; however, ground-truth punctuation is not present in speech recognition output.  

\paragraph{Structured prediction as sequence-to-sequence.}
\citet{NIPS2015_277281aa} show that attention-enhanced sequence-to-sequence models can be trained for complex structured prediction tasks such as syntactic parsing.
\citet{t5} takes a step further to model all text-based language problems in a text-to-text format.
\citet{paolini2021structured} framed many NLP tasks as translation between augmented natural languages.

\paragraph{Constrained decoding.}
\citet{hokamp-liu-2017-lexically} and \citet{post-vilar-2018-fast} introduced lexical constraints in neural machine translation beam search.
\citet{anderson-etal-2017-guided} formulated lexical constraints as finite-state machines.
\citet{deutsch-etal-2019-general} used an active set method to efficiently compose many automata with beam search.




\section{Conclusion}

We have presented new methods for long-form speech translation by coupling source-side large language models with finite-state decoding constraints, allowing large language models to be used for structured prediction
with a guarantee for wellformedness in the output space.
Finite-state constraints are especially effective when the model is decoder-only, relatively small, or has not been completely fine-tuned (only prompt-tuned, or few-shot-learned) for the structured prediction task. We also observe that even though complete fine-tuning and enlarging model size can reduced the rate of invalid output, models alone are not capable of completely eliminate invalid output. 

Fine-tuning on in-domain ASR transcripts containing recognition errors and disfluency improves segmentation accuracy over training on clean transcripts. Our qualitative analysis shows the largest category of remaining errors is ASR errors which make transcripts difficult to parse and segment.
The fact that LLMs are capable of adapting to ASR errors points to future research directions of contextualized ASR error recovery.




\makeatletter
\ifacl@finalcopy
    
    \section*{Acknowledgments}
    
    We thank Chu-Cheng Lin and Nicholas Tomlin for comments that improved the presentation of the work.
    A.D.M. is supported by an Amazon Fellowship and a Frederick Jelinek Fellowship.
    
\fi
\makeatother


\section*{Limitations}
Large language models are more expensive and slower compared to dedicated smaller models for sentence segmentation. The additional latency introduced in the speech-to-text cascade by such models can be too high for online processing.

We use a sliding window algorithm to combine segmentation outputs from adjacent fixed-size  text windows. The choice is a sub-optimal heuristic efficiency-accuracy tradeoff. Recently, large language models have become increasingly capable of handling long paragraphs. We can simplify the system by applying large language models directly on long paragraphs. However, as the output length increases, the likelihood of hallucinations increases, making decoding constraints more important. Moreover, there may always be long-form audio whose transcriptions exceed the context length of even the largest language models.

Finally, adhering to the cascade architecture---speech-recognition followed by text-to-text translation---introduces the problem of error propagation. Our error analysis has shown that speech recognition errors form the main category within the remaining errors made by our systems. Can text-only large language models systematically correct speech recognition errors without introducing hallucinations? Furthermore, can a speech recognition model that incorporates a large language model jointly recognize and segment the transcription better than a cascade system?








\bibliography{custom}
\bibliographystyle{acl_natbib}

\end{document}